\documentclass[conference]{IEEEtran}
\usepackage[T1]{fontenc}
\usepackage{amsmath,amssymb,graphicx,booktabs,longtable}
\usepackage{cite}
\usepackage[hidelinks]{hyperref}
\usepackage{url}
\begin{document}
\title{ShiftGuard: Evaluating Confidence and Feature Distance for Risk-Controlled Image Classification}
\author{\IEEEauthorblockN{Mitra Varun Gogulapati}\IEEEauthorblockA{\textit{Independent Researcher}\\mitravarun2702@gmail.com}}
\maketitle

\begin{abstract}
A classifier that is usually right still makes confident mistakes, and a deployed system needs some way to decide when to answer and when to abstain. Cheap post-hoc scores such as maximum softmax probability (MSP) and predictive entropy are the standard answer. Distances in feature space are a popular alternative, and fusing several signals into one score is a natural next step. We ask whether adding a feature-distance signal to MSP and entropy improves error detection, and whether it increases the share of predictions that can be accepted at a certified error rate when only a small labelled calibration set is available. We train a CIFAR-adapted ResNet-18 from scratch on CIFAR-10 (92.92\% test accuracy, one seed) and compare five reliability scores: MSP, entropy, distance to the predicted-class centroid, a logistic fusion of MSP and entropy, and a logistic fusion of all three signals. Acceptance thresholds are chosen on a held-out calibration set with Clopper-Pearson upper bounds, Bonferroni-corrected over a prespecified family of 315 rules at 95\% simultaneous confidence. Confidence-based scores were best. Error AUROC was 0.9230 for MSP, 0.9233 for entropy and 0.9235 for MSP plus entropy, against 0.8814 for the three-signal fusion and 0.8224 for distance alone. At a 5\% risk target with 1,000 calibration images, entropy accepted 84.82\% of test images with 2.05\% observed error, while three-signal fusion accepted 80.61\% with 2.21\%. With 250 calibration images, neither distance nor the three-signal fusion could certify the 5\% target at all. The results come from one seed, one dataset and one architecture, with no corruption experiments, so they say nothing about distribution-shift detection. Within these limits, the distance signal did not earn its added complexity.
\end{abstract}
\section{Introduction}

Image classifiers are usually judged by accuracy, but accuracy hides a practical problem. A network that is right 93\% of the time is wrong 7\% of the time, and when it is wrong it is often very sure of itself \cite{r18}. A system built around such a classifier needs a second decision on top of the prediction: accept it, or hand the image to a person or a slower model. This is selective classification, and its roots go back to the reject option of Chow \cite{r7} and the formal treatment of El-Yaniv and Wiener \cite{r11}. Geifman and El-Yaniv \cite{r15} brought it to deep networks and showed how to choose a rejection threshold with a statistical guarantee on the error rate of the accepted predictions.

The cheapest way to build a rejection rule is to reuse something the classifier already computes. The maximum softmax probability is the usual baseline for spotting errors \cite{r22}, and the entropy of the softmax distribution is a close relative. Both look only at the output layer. A different family of methods looks inside the network instead. These ask how far an input's learned representation sits from the representations of examples the model handled correctly, with Trust Score \cite{r24}, Mahalanobis distance \cite{r27} and nearest-neighbour distance \cite{r36} as well-known examples. The two families see different things. It is therefore tempting to combine them, for instance by fitting a small logistic regression on the signals. If the signals carry complementary information, the fused score should rank errors better and should let more predictions through at the same error rate.

Whether that actually happens, and whether it still happens when the labelled data available for calibration are scarce, is the question this paper asks. Scarcity matters because a certificate is only as strong as the data behind it. With a few hundred labelled calibration images, the statistical upper bound on accepted-set risk is wide, and a rule that looks good on the calibration data may not be provably good. A method with worse error ordering can lose certified coverage in this regime. All five scores receive the same per-rule confidence adjustment; the bound does not directly penalise the number of fitted logistic coefficients.

We built a small evaluation framework, which we call ShiftGuard, to answer it. A CIFAR-adapted ResNet-18 is trained from scratch, and a reliability layer on top produces five scores from the frozen classifier. Thresholds are chosen on separate calibration data and certified with finite-sample bounds, and everything is evaluated once on the official test set. The name reflects the original motivation of guarding a classifier against distribution shift. The experiments in this paper use clean, in-distribution data only, so the name should not be read as a claim of demonstrated shift detection. We return to this in Section 7.

Our research question is the following.

Does adding a feature-distance signal to confidence and entropy improve error detection, and does it increase coverage at a statistically controlled error rate, particularly when calibration data are limited?

Our contributions are these.

\begin{enumerate}
\item A controlled comparison of five post-hoc reliability scores on a single frozen ResNet-18, with disjoint data for training, checkpoint selection, class-centroid estimation, fusion fitting, threshold calibration and testing.
\item A risk-control audit that keeps three quantities apart: the target risk, the calibration upper bound used to certify a rule, and the error rate observed on the test set. It covers two risk targets (5\% and 10\%) and three calibration budgets (250, 500 and 1,000 images), with one simultaneous confidence level over all 315 prespecified rules.
\item An empirical answer that is negative for fusion. Confidence-based scores matched or beat both distance-based variants on every ranking metric and on certified coverage, and the three-signal fusion could not certify the 5\% target with 250 calibration images and had substantially lower coverage with 500.
\item An auditable saved-output package with an independent script for score reconstruction, certificate verification and paired bootstrap analysis.
\end{enumerate}

We make no novelty claim for the statistical machinery. Confidence fusion is common, and the binomial bound with a Bonferroni correction is an established construction. The contribution is the empirical comparison.

The paper is organised as follows. Section 2 covers related work. Section 3 describes the scores, the fusion and the certification procedure. Section 4 gives the experimental setup, Section 5 the results, Section 6 a discussion, Section 7 the limitations and Section 8 the conclusion. The appendix holds complete certification tables, extra figures and exploratory earlier experiments.

\section{Related Work}

Selective classification and failure prediction. The reject option was studied for pattern recognition by Chow \cite{r7}, who characterised the trade-off between error and rejection rates. El-Yaniv and Wiener \cite{r11} gave a theoretical framework for selective classification with a risk-coverage curve. Geifman and El-Yaniv \cite{r15} applied it to deep networks, ranked inputs by softmax confidence, and proposed an algorithm that picks a threshold with a guaranteed bound on selective risk using a binomial confidence interval. The area under the risk-coverage curve (AURC) was used as a summary metric in later work on estimating and improving selective risk \cite{r17}. SelectiveNet \cite{r16} trains the abstention decision into the network. For ranking errors without a rejection threshold, Hendrycks and Gimpel \cite{r22} showed that MSP is a strong baseline for detecting misclassified inputs, and later work learns a separate confidence network for failure prediction \cite{r9}. Large comparisons of uncertainty estimators on image classifiers keep finding that simple confidence scores are hard to beat once classifiers are trained well \cite{r14,r3}. Our study is a small, in-distribution instance of that question with a distance signal added.

Feature-space and distance-based scores. Several methods score inputs by their position in representation space. Trust Score \cite{r24} compares the distance to the predicted class's high-density region with the distance to the nearest other class. The Mahalanobis detector \cite{r27} fits class-conditional Gaussians with a shared covariance on penultimate features and scores inputs by distance to the nearest class mean, and nearest-neighbour methods use distances to training examples \cite{r36}. Energy scores \cite{r28} come from the logits and sit between the two families. These methods were mostly developed and evaluated for out-of-distribution detection. Our distance score is deliberately simpler. It is the root-mean-square Euclidean distance, in per-dimension standardised coordinates, from an input's feature vector to the centroid of its predicted class. It ignores correlations between feature dimensions and class multimodality, and it is not Mahalanobis distance or Trust Score. We chose it to isolate whether a basic geometric signal adds anything, and we do not claim it represents the best that distance-based methods can do.

Combining uncertainty signals. Ensembles \cite{r26} and Monte Carlo dropout \cite{r13} improve uncertainty by averaging multiple predictions, at the price of extra forward passes. Cheaper alternatives fit a small model on top of existing signals, which is the approach we take with logistic regression. The logistic combination optimises likelihood of the error labels, which is a different objective from ranking, and we will see in Section 6 that this difference may matter. Benchmarks of post-hoc selective-classification methods report that outcomes depend on the classifier and the signal \cite{r6}, which is one reason we report the full comparison and not only the final fused score.

Calibration. Modern networks tend to be overconfident \cite{r18}. Expected calibration error (ECE) bins predictions by confidence and averages the gap between confidence and accuracy \cite{r31}, and the Brier score measures squared probability error \cite{r5}. We report these for the classifier and also for the error probabilities that the fused scores produce, because ranking quality and probability calibration are different properties.

Statistical risk control. Distribution-free methods choose a decision rule so that a risk is bounded with high probability over the calibration sample. Risk-controlling prediction sets \cite{r4} and Learn then Test \cite{r2} treat rule selection as multiple hypothesis testing over a grid of candidate parameters. Conformal prediction is the best-known relative \cite{r1}. The version we use is the simplest member of this family. For each candidate threshold on a grid fixed in advance, we compute a one-sided Clopper-Pearson upper bound \cite{r8} on the accepted-set error rate and correct for the size of the grid with a Bonferroni adjustment \cite{r10}. This is the same idea as the guaranteed-risk selective classification of Geifman and El-Yaniv \cite{r15}, extended over several methods and budgets.

Evaluation under shift. Predictive uncertainty can degrade sharply under dataset shift \cite{r32}, and corrupted-image benchmarks exist for measuring robustness \cite{r21}. Even without deliberate corruption, a new test set drawn by a nominally identical process can shift accuracy \cite{r35}. These results are why we state the IID assumption behind our certificates explicitly, and why we do not extend the claims to shifted data.

\section{Methodology}

\subsection{Problem setup}

A trained classifier f maps an image x to logits z(x) $\in$ $\mathbb R^{10}$ and a penultimate feature vector h(x) $\in$ $\mathbb R^{512}$. The predicted class is

\begin{equation}
\hat y(x) = \arg\max_k z_k(x)
\end{equation}

with softmax probabilities $p_k(x)=\exp(z_k(x))/\sum_j\exp(z_j(x))$. Define $e(x,y)=\mathbf1\{\widehat y(x)\ne y\}$. The classifier is frozen after training; reliability scores do not retrain it.

A reliability score s(x) is oriented so that higher means more likely to be wrong. A selective classifier accepts the prediction when s(x) $\le$ $\tau$ for a threshold $\tau$ and abstains otherwise. For a set of N examples, the coverage of a threshold is the fraction accepted, and the selective risk is the error rate among accepted examples:

\begin{equation}
\begin{aligned}\widehat C(\tau)&=N^{-1}\sum_{i=1}^{N}\mathbf1\{s_i\le\tau\},\\\widehat R(\tau)&=\frac{\sum_{i:s_i\le\tau} e_i}{\sum_i\mathbf1\{s_i\le\tau\}}.\end{aligned}
\end{equation}

When no example is accepted, R($\tau$) is undefined. We never replace it with zero.

\subsection{The five reliability scores}

MSP. The score is one minus the maximum softmax probability,

\begin{equation}s_{\rm MSP}(x)=1-\max_k p_k(x).\end{equation}

so lower confidence gives a higher score.

Entropy. The score is the Shannon entropy of the softmax distribution in natural logarithms, with probabilities clipped at $10^{-15}$ inside the logarithm for numerical stability,

\begin{equation}s_{\rm ent}(x)=-\sum_{k=1}^{10}p_k(x)\log p_k(x).\end{equation}

Feature distance. Only correctly classified examples in an independent 1,000-image reference set are used. Per-dimension means $\mu$ and standard deviations $\sigma$ are estimated from their features, with SD floored at 0.001. Set $u(x)=(h(x)-\mu)/\sigma$, with elementwise division. The centroid $c_k$ averages standardised features of correct reference examples of class $k$. The score is the RMS distance to the predicted-class centroid:

\begin{equation}s_{\rm dist}(x)=\sqrt{\frac1{512}\|u(x)-c_{\widehat y(x)}\|_2^2}.\end{equation}

A class with fewer than two correct reference examples would fall back to the pooled correct-example centroid. This did not occur here, since every class had between 86 and 98 correct reference examples. We keep the reference set disjoint from fusion fitting to avoid optimistic in-sample distances. Independent calibration of a fixed score is essential for the certificate.

MSP + entropy. A logistic regression is fitted on the standardised pair (MSP, entropy), with the error indicator as the label. The fitted score is the predicted error probability.

All three signals. The same logistic regression is fitted on the standardised triple (MSP, entropy, distance).

For both fused scores we use a standard scaler followed by unweighted logistic regression with L2 regularisation strength C = 1, fixed in advance and not tuned. The estimator is fitted on a 1,000-image fusion-fitting set that is disjoint from the reference set, the calibration set and the test set. The fitted weights are stored. In the three-signal model the standardised coefficients were 0.767 (MSP), 1.558 (entropy) and 0.588 (distance), with intercept -3.146. In the two-signal model they were 0.836 and 1.804 with intercept -2.977.

\paragraph{Score convention audit.}
Let $q=\max_k p_k$. The standalone MSP error score is $1-q$, but the first fusion input is $q$ itself. This is verified by the saved input matrix and by reconstructing both fitted probabilities from the exported coefficients to numerical precision. For $v=(q,H,d)$ and fitting-set means $a$ and scales $b$, the fitted probability is
\begin{equation}
\widehat P(e=1\mid v)=\operatorname{sigmoid}\left(\beta_0+\sum_j\beta_j\frac{v_j-a_j}{b_j}\right).
\end{equation}
The two-signal model omits $d$. Reparameterising the first input as $1-q$ requires negating its coefficient and replacing its mean by $1-a_1$; it does not change predictions. The positive recorded coefficient on $q$ is conditional on the other signals and is not a standalone confidence-to-error relationship.

\subsection{Error-detection and selective-prediction metrics}

Error AUROC treats an incorrect prediction as the positive class and measures how well the score separates errors from correct predictions \cite{r12}. Ties in the score do not affect it beyond the standard rank treatment.

Accepting in increasing score order gives $s_{(1)}\le\cdots\le s_{(N)}$. The prefix risk is $R_k=k^{-1}\sum_{i=1}^{k}e_{(i)}$. We use the discrete area under the risk--coverage curve:

\begin{equation}
\text{AURC} = \frac{1}{N}\sum_{k=1}^{N} R_k
\end{equation}

Lower is better. Groups of exactly tied scores use the expected prefix risk under a random ordering within the group, so that the result does not depend on the order in which tied examples happen to appear. We also report selective risk at fixed coverages of 50\%, 70\%, 80\% and 90\%, computed at ceil(coverage $\times$ N) accepted examples.

Calibration of the classifier uses the negative log-likelihood of the true class (probabilities clipped to [$10^{-15}$, 1]), the multiclass Brier score, and ECE with 15 equal-width confidence bins. For the fused scores, which output error probabilities, we report the binary Brier score and a binary ECE on the same 15-bin scheme. These describe whether the numbers can be read as probabilities, which is separate from how well they rank.

\subsection{Risk-controlled threshold selection}

Ranking metrics do not choose a threshold. To do that with a guarantee, we use independent calibration data. The procedure has four steps.

Step 1, candidate thresholds. For each method, the candidate thresholds are fixed before any calibration label is used. They are the unique values among 21 equally spaced quantiles (0, 0.05, ..., 1) of that method's scores on the fusion-fitting set. The calibration and test sets play no part in forming the grid.

Step 2, bounds. For a calibration budget n, the first n calibration images are used. A threshold $\tau$ accepts m examples of which k are errors. The one-sided Clopper-Pearson upper confidence bound on the accepted-set risk is the (1 - $\delta$') quantile of a Beta(k + 1, m - k) distribution,

\begin{equation}U(k,m,\delta\prime)=\operatorname{Beta}^{-1}(1-\delta\prime;k+1,m-k).\end{equation}

and U = 1 when m = 0 or k = m. The per-rule level is $\delta$' = $\delta$ / |F|, where |F| is the number of rules in the prespecified family and $\delta$ = 0.05.

Step 3, family size. The family contains every combination of method, candidate threshold and calibration budget. With 5 methods, 21 candidate thresholds each and 3 budgets, |F| = 315, giving $\delta$' = 0.05/315 $\approx$ 1.59 $\times$ $10^{-4}$. The two risk targets add no further tests, because the same upper-bound event is used for both.

Step 4, selection. For a method, budget and target $\alpha$, a rule is eligible if it accepts at least 20 calibration examples and U $\le$ $\alpha$. Among eligible rules, the one with the largest calibration acceptance count is chosen, with ties broken by the smaller upper bound and then the smaller threshold. If no rule is eligible, the method returns "not certified", accepts no test examples, and its conditional error rate is reported as undefined and never as zero.

The guarantee that follows is this. If the calibration and future clean examples are IID draws from one population, and the model, the scores and the candidate grids are fixed before calibration labels are looked at, then with probability at least 0.95 over the calibration sample, every rule in the family has a population accepted-set error rate no larger than its computed bound U. A selected rule therefore has population risk at most its bound, which is at most $\alpha$. The guarantee concerns a population rate under stated assumptions. It does not apply to individual images, to the particular test sample, or to arbitrary distribution shift. Because the bound holds simultaneously over a fixed finite family, it remains valid even though the selection step looks across the whole grid.

The calibration budgets 250, 500 and 1,000 are nested prefixes of the same 1,000 calibration images. The budgets are therefore evaluations of one sample at different sizes, and results across budgets are not independent replications.

\subsection{Three quantities that must stay separate}

We use three different risk numbers and they should never be read as one.

\noindent The target risk $\alpha$ is the requirement we set, 5\% or 10\%.

\noindent The calibration upper bound U is a conservative statistical bound computed from calibration data, and it is what makes a rule eligible.

\noindent The observed test risk is the actual fraction of accepted test predictions that were wrong. It is a measurement on one finite sample and fluctuates around the true population risk. It can be above or below the target. It is not what certifies a rule.

Because U is deliberately conservative, observed test risk is typically far below $\alpha$, as Section 5.4 shows.

\begin{figure*}[!t]
\centering
\includegraphics[width=\textwidth,height=.33\textheight,keepaspectratio]{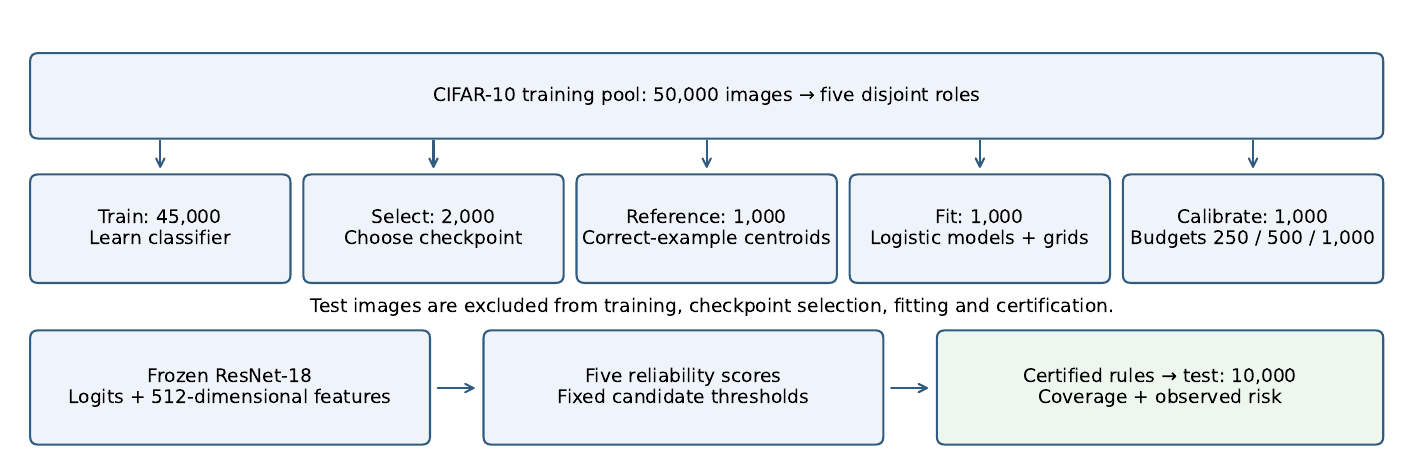}
\caption{Data roles and information flow. Scores come from a frozen classifier; threshold grids use fitting data, certificates use calibration data, and final outcomes use the official test set.}\label{fig:pipeline}
\end{figure*}

\section{Experimental Setup}

\subsection{Dataset and data roles}

We use CIFAR-10 \cite{r25}, which has 32 $\times$ 32 colour images in ten classes. The archive was downloaded from a pinned mirror revision and checked against the official archive's MD5 hash (Appendix B). The official 50,000-image training set is divided into five disjoint subsets, and the official 10,000-image test set is kept apart.

\begin{table*}[!t]
\centering\footnotesize\setlength{\tabcolsep}{4pt}
\caption{Disjoint data roles. The first five rows partition the official training pool.}
\begin{tabular}{lrr}\hline
Role & Images & Purpose\\\hline
Training & 45,000 & Classifier weights\\
Selection & 2,000 & Checkpoint epoch\\
Reference & 1,000 & Correct-example centroids\\
Fusion fitting & 1,000 & Estimator + threshold grid\\
Risk calibration & 1,000 & Threshold certificates\\
Official test & 10,000 & Final evaluation\\
\hline\end{tabular}
\end{table*}

The risk-calibration subset was set aside without using labels before the remaining pool was stratified, so its class counts are not exactly balanced (they range from 83 to 118 per class). The training, selection, reference and fitting subsets are stratified by class. The reference set holds 931 correctly classified examples (69 errors), and the fitting and calibration sets contain 76 and 81 classifier errors, respectively. The test set contains 708 errors. No test label was used to train, to select a checkpoint, to fit a score, to set a candidate grid or to choose a threshold.

\subsection{Classifier and training}

The classifier is a ResNet-18 \cite{r20} trained from random initialisation with no pretrained weights. For 32 $\times$ 32 inputs we replace the first convolution with a 3 $\times$ 3, stride-1 convolution with 64 filters and remove the initial max-pooling layer. The remaining residual stages follow the standard torchvision ResNet-18. The model has 11,173,962 parameters and outputs a 512-dimensional feature vector after global average pooling, followed by a linear layer with ten outputs.

Training uses AdamW \cite{r30} with an initial learning rate of 0.001, weight decay 0.0001 and batch size 128, for 40 epochs. The learning rate follows cosine annealing \cite{r29} down to a minimum of 0.00001. Training images receive random 32-pixel crops with 4-pixel padding and random horizontal flips. All images are normalised with the standard CIFAR-10 channel means (0.4914, 0.4822, 0.4465) and standard deviations (0.2470, 0.2435, 0.2616). Selection, reference, fitting, calibration and test images use deterministic preprocessing without augmentation. The checkpoint is the epoch with the highest accuracy on the 2,000-image selection set. That was epoch 39, with a selection accuracy of 93.95\%. The seed was 42.

\begin{table*}[!t]
\centering\footnotesize\setlength{\tabcolsep}{4pt}
\caption{Prespecified main-run configuration.}
\begin{tabular}{lr}\hline
Setting & Value\\\hline
Architecture & CIFAR-adapted ResNet-18\\
Training & 40 epochs; seed 42; batch 128\\
Optimiser & AdamW; weight decay 0.0001\\
Learning rate & 0.001; cosine minimum 0.00001\\
Fusion & StandardScaler + logistic; C = 1\\
Candidate family & 5 methods; 21 thresholds; 3 budgets\\
Calibration budgets & 250 / 500 / 1,000\\
Risk targets & 5\% / 10\%\\
Confidence & 95\% simultaneous over 315 rules\\
Acceptance floor & 20 calibration examples\\
\hline\end{tabular}
\end{table*}

Additional diagnostics are provided in Appendix D.

\subsection{Software, hardware and cost}

The run session records Google Colab execution with a Tesla T4 GPU and Python 3.13.16. The recorded environment was PyTorch 2.11.0 (CUDA 13.0), torchvision 0.26.0, NumPy 2.1.3, scikit-learn 1.6.1, SciPy 1.16.3 and Matplotlib 3.10.0 \cite{r33,r19,r34,r37,r23}. Training took 2,837 seconds (about 47 minutes, around 71 seconds per epoch), not counting the time to upload the final checkpoint. Peak process memory was 1,928 MiB and peak CUDA allocation 877 MiB. On the test set, the forward pass took 0.268 ms per image and feature extraction 6.2 seconds in total, and computing a combined score took 0.008 ms per image. The reliability layer therefore adds almost nothing to inference cost. Analysis ran in float32 or float64 as recorded in the code.

The Colab workflow saves checkpoints and intermediate results to persistent storage and resumes interrupted training from the last completed epoch. The draft reports a 20-test suite and a prior CPU smoke run. The supplied results archive contains no test source or execution log, so those development checks are not independently verified here. The saved scores, disjoint split indices, coefficients, bounds and selected-rule outcomes were independently audited for this manuscript.

\subsection{Protocol integrity}

Candidate threshold grids come from fusion-fitting scores only. Calibration data are used only to compute bounds and select a rule from the fixed grid. The reported protocol reserves test data until the classifier, fusion models, grids and certificates are frozen. The results archive supports checking saved artefacts, but cannot independently establish every aspect of execution order. The repository records SHA-256 hashes of the checkpoint and the configuration so that an evaluation can be checked against the exact model and settings (Code and Data Availability).

\section{Results}

All results in this section come from one training seed (42) and the official 10,000-image test set. No across-seed intervals are available, and we say so wherever a comparison is close.

\subsection{Classifier performance}

\begin{table}[!t]
\centering\footnotesize\setlength{\tabcolsep}{4pt}
\caption{Frozen classifier performance on the official test set.}
\begin{tabular}{lr}\hline
Metric & Value\\\hline
Accuracy & 92.92\%\\
Correct / incorrect & 9,292 / 708\\
NLL & 0.3789\\
Multiclass Brier & 0.1202\\
15-bin ECE & 0.0488\\
\hline\end{tabular}
\end{table}

Additional diagnostics are provided in Appendix D.

Additional diagnostics are provided in Appendix D.

The classifier is accurate enough to make a meaningful reliability study. Its ECE of 0.0488 is modest but not negligible. On the 1,000-image fitting set, the mean maximum softmax probability was 0.981 while accuracy was 92.4\%, so the network is overconfident in the usual way. The confusion matrix shows that errors are concentrated in a few animal classes. The largest single confusions are dog predicted as cat (67 images) and cat predicted as dog (66 images), followed by cat predicted as bird (28), bird as deer (26) and cat as deer (25). Per-class recall is lowest for cat (83.7\%) and dog (89.6\%) and highest for automobile (97.1\%) (Appendix D).

\subsection{Error detection}

\begin{table}[!t]
\centering\footnotesize\setlength{\tabcolsep}{4pt}
\caption{Error detection: higher AUROC and lower AURC are better.}
\begin{tabular}{lrr}\hline
Method & AUROC & AURC\\\hline
MSP & 0.92299 & 0.00860\\
Entropy & 0.92332 & 0.00857\\
Distance & 0.82242 & 0.01845\\
MSP + entropy & 0.92354 & 0.00856\\
All three & 0.88145 & 0.01357\\
\hline\end{tabular}
\end{table}

\begin{figure*}[!t]
\centering
\includegraphics[width=\textwidth,height=.33\textheight,keepaspectratio]{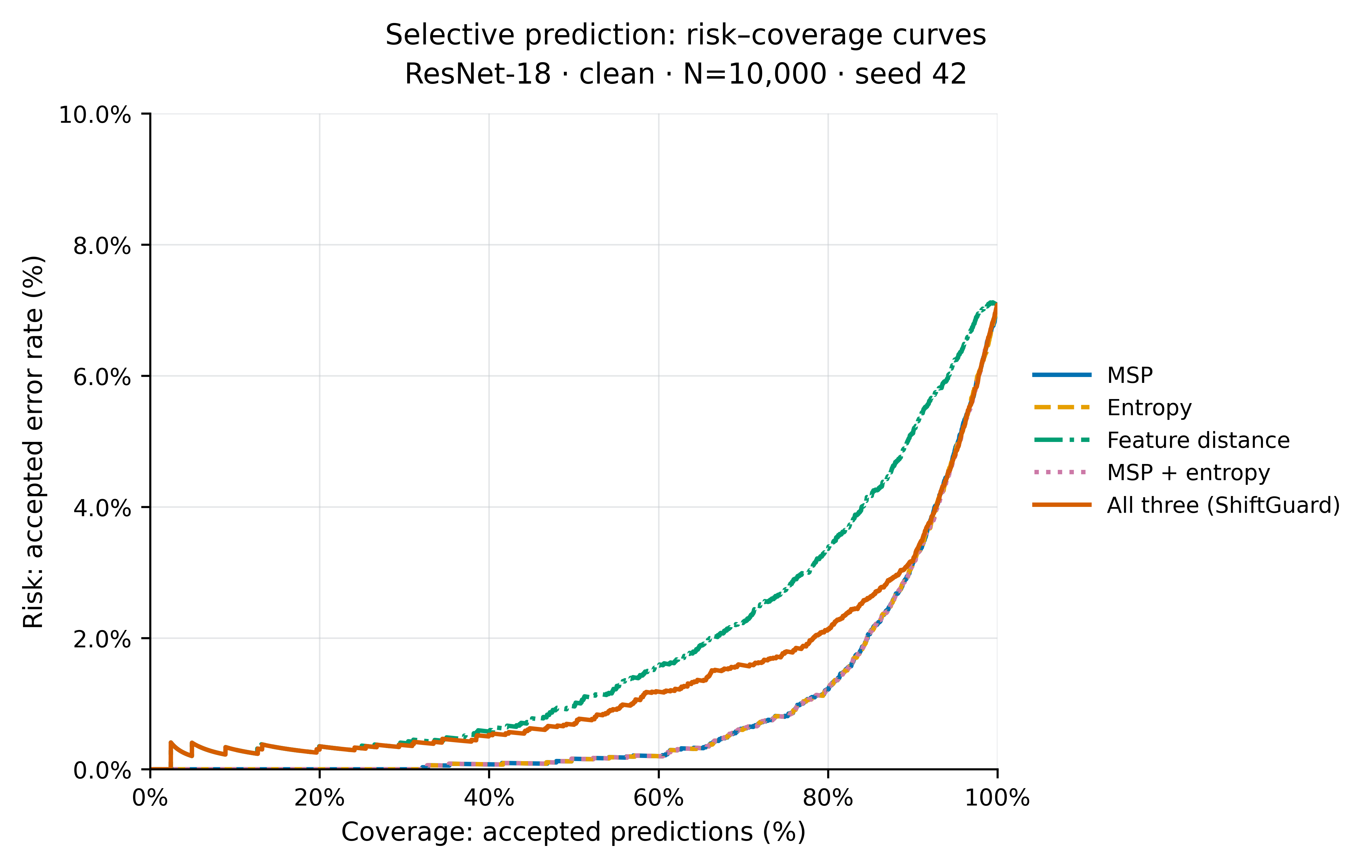}
\caption{Risk-coverage curves on 10,000 clean test images. Lower is better. AURC averages prefix risks with expected random-order treatment of exact ties. One training seed.}\label{fig:risk}
\end{figure*}

Three confidence-based scores are nearly identical. MSP, entropy and the two-signal fusion differ in the third decimal place of AUROC and in the fifth of AURC. That is expected, because for a ten-class softmax output MSP and entropy are strongly related, and their rank correlation on the test set rounds to 1.000. The two-signal fusion has the best numbers (0.92354 and 0.00856), but its edge over MSP or entropy alone is tiny, and Section 5.7 shows that its paired interval includes zero.

The distance signal alone is clearly worse, with an AUROC of 0.822 against 0.923 for MSP. It is not uninformative. The mean standardised distance was 0.558 for correctly classified test images and 0.714 for misclassified ones, so errors do sit farther from their predicted centroid on average. But as a ranking of errors it is weaker than confidence.

Adding distance to the two-signal fusion made things worse, not better. The three-signal score has an AUROC of 0.881, which is 0.042 below the two-signal fusion, and an AURC of 0.01357 against 0.00856, an increase of about 59\%. It sits between the distance signal and the confidence signals. This is the central result of the paper: in this run, fusing the feature-distance signal with confidence and entropy lowered error-detection quality.

\subsection{Selective risk at fixed coverage}

\begin{table*}[!t]
\centering\footnotesize\setlength{\tabcolsep}{4pt}
\caption{Observed test risk in percent (error count) at fixed coverage.}
\begin{tabular}{lrrrr}\hline
Method & 50\% & 70\% & 80\% & 90\%\\\hline
MSP & 0.16 (8) & 0.61 (43) & 1.23 (98) & 3.14 (283)\\
Entropy & 0.16 (8) & 0.61 (43) & 1.23 (98) & 3.12 (281)\\
Distance & 0.96 (48) & 2.24 (157) & 3.36 (269) & 5.12 (461)\\
MSP + entropy & 0.16 (8) & 0.61 (43) & 1.24 (99) & 3.13 (282)\\
All three & 0.68 (34) & 1.59 (111) & 2.14 (171) & 3.18 (286)\\
\hline\end{tabular}
\end{table*}

\begin{figure*}[!t]
\centering
\includegraphics[width=\textwidth,height=.33\textheight,keepaspectratio]{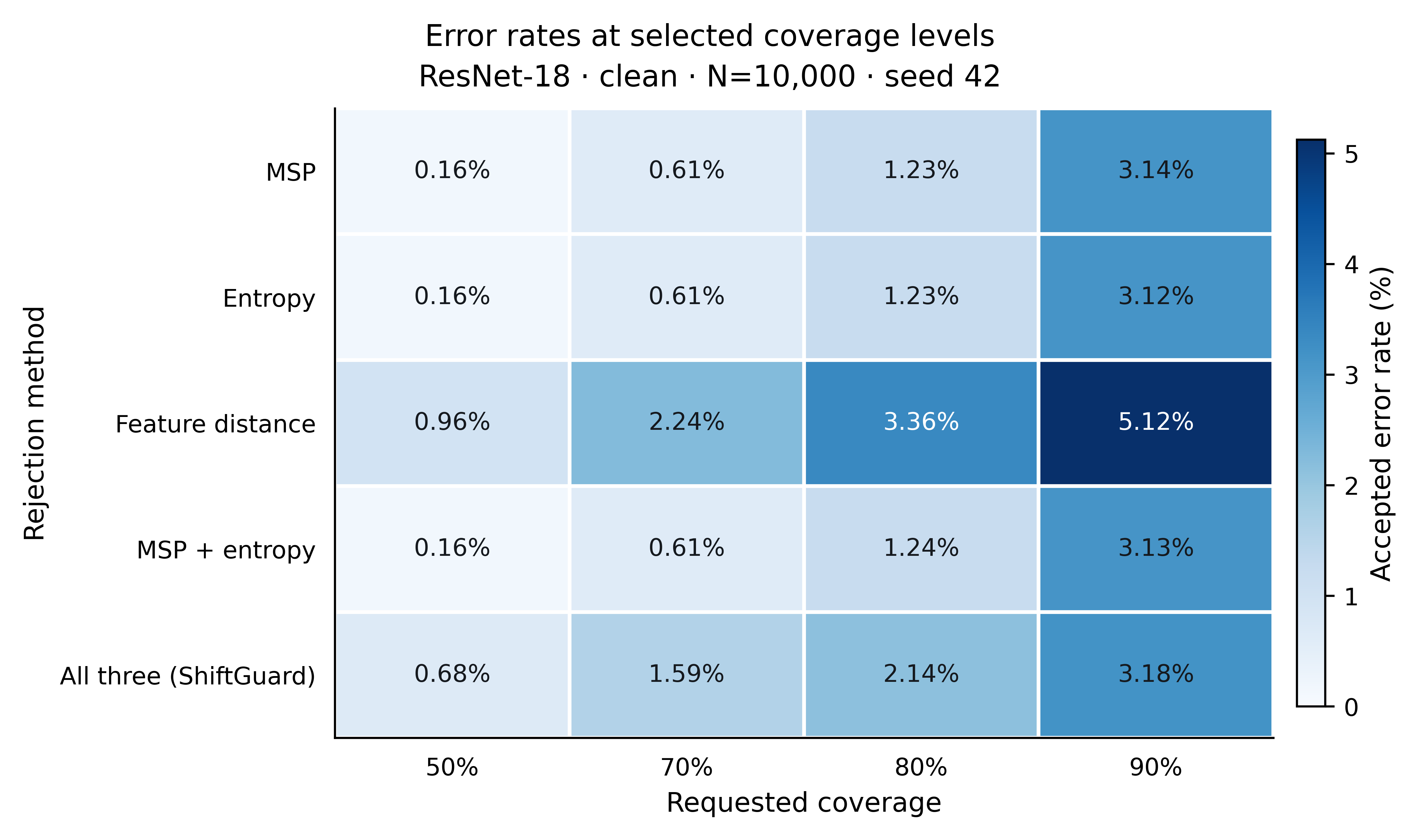}
\caption{Observed test risk (\%) at fixed coverage, with identical test images for every method. One seed; no population guarantee is attached to these fixed-coverage summaries.}\label{fig:fixed}
\end{figure*}

The count of errors makes the gaps concrete. When the 5,000 most trusted predictions are accepted, entropy lets through 8 errors, while the three-signal fusion lets through 34 and distance alone 48. Put the other way, entropy places 700 of the 708 errors in the least trusted half of the data. At 70\% coverage the counts are 43 for entropy and 111 for the three-signal score, and at 80\% they are 98 and 171. The gap closes at 90\% coverage (281 versus 286 errors). So the extra signal costs the most at the top of the ranking, where a few confident mistakes are what a deployment would most like to avoid, and costs almost nothing once most of the data are accepted. At 90\% coverage, rejecting the 1,000 least trusted images removes 425 of the 708 errors under MSP (60\%) and 422 under the three-signal score.

\subsection{Controlled-risk coverage}

\begin{table*}[!t]
\centering\footnotesize\setlength{\tabcolsep}{4pt}
\caption{Certification audit with 1,000 calibration images. U is the calibration upper bound; test risk is descriptive.}
\begin{tabular}{lrrrrrr}\hline
Method & Target & Cal. accepted/errors & U & Test accepted/errors & Coverage & Test risk\\\hline
MSP & 5\% & 845/15 & 4.04\% & 8480/175 & 84.80\% & 2.06\%\\
MSP & 10\% & 935/47 & 8.10\% & 9365/408 & 93.65\% & 4.36\%\\
Entropy & 5\% & 846/15 & 4.03\% & 8482/174 & 84.82\% & 2.05\%\\
Entropy & 10\% & 938/50 & 8.47\% & 9397/418 & 93.97\% & 4.45\%\\
Distance & 5\% & 622/11 & 4.53\% & 6196/102 & 61.96\% & 1.65\%\\
Distance & 10\% & 887/57 & 9.90\% & 8933/445 & 89.33\% & 4.98\%\\
MSP + entropy & 5\% & 846/15 & 4.03\% & 8479/174 & 84.79\% & 2.05\%\\
MSP + entropy & 10\% & 936/49 & 8.36\% & 9388/407 & 93.88\% & 4.34\%\\
All three & 5\% & 799/17 & 4.62\% & 8061/178 & 80.61\% & 2.21\%\\
All three & 10\% & 929/47 & 8.16\% & 9375/409 & 93.75\% & 4.36\%\\
\hline\end{tabular}
\end{table*}

\begin{figure*}[!t]
\centering
\includegraphics[width=\textwidth,height=.33\textheight,keepaspectratio]{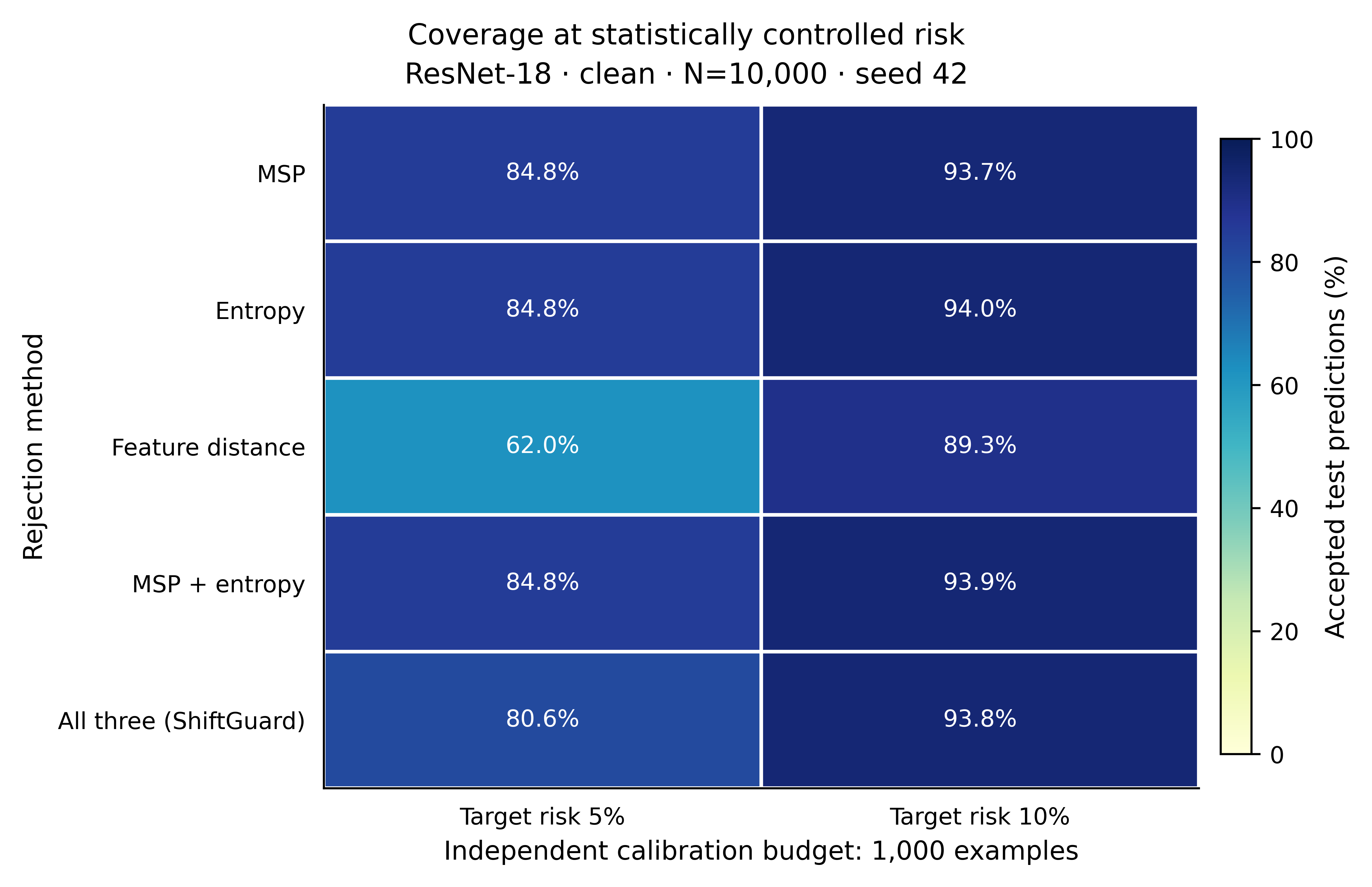}
\caption{Test coverage of certified rules. One simultaneous 315-rule family covers all methods, thresholds and budgets. Not-certified rules reject every example; their conditional risk is undefined. Validity assumes IID clean data.}\label{fig:cert}
\end{figure*}

With the full 1,000-image calibration set and a 5\% target, entropy accepted 8,482 of the 10,000 test images (84.82\%) and 174 of those were wrong, an observed accepted-set error rate of 2.05\%. The three-signal fusion accepted fewer, 8,061 (80.61\%), with 178 errors (2.21\%). Distance alone was the most conservative at 61.96\% coverage. MSP and the two-signal fusion matched entropy to within a few images. At the 10\% target, entropy accepted 93.97\% of images at 4.45\% observed error, and the three-signal fusion accepted 93.75\% at 4.36\%, which is a very small difference.

Two things are worth stating plainly. First, the observed test risk is well below the target in every row, around 2\% for a 5\% target and around 4.4\% for a 10\% target. The upper bounds are conservative by construction, since they must hold simultaneously for 315 rules at 95\% confidence. Second, the 10\% target is not demanding for this classifier. The overall test error rate is 7.08\% (708/10,000), so accepting every prediction would already meet it on this sample. The 5\% target is the informative one, and it is where the methods separate.

\subsection{Effect of the calibration budget}

\begin{table*}[!t]
\centering\footnotesize\setlength{\tabcolsep}{4pt}
\caption{Certified-rule test coverage (\%); n.c. denotes not certified and undefined conditional risk.}
\begin{tabular}{lrrrr}\hline
Method & Target & 250 & 500 & 1,000\\\hline
MSP & 5\% & 70.79 & 80.36 & 84.80\\
Entropy & 5\% & 70.84 & 80.56 & 84.82\\
Distance & 5\% & n.c. & n.c. & 61.96\\
MSP + entropy & 5\% & 70.84 & 80.55 & 84.79\\
All three & 5\% & n.c. & 67.67 & 80.61\\
MSP & 10\% & 88.61 & 93.65 & 93.65\\
Entropy & 10\% & 88.56 & 93.97 & 93.97\\
Distance & 10\% & 71.82 & 84.98 & 89.33\\
MSP + entropy & 10\% & 88.54 & 93.88 & 93.88\\
All three & 10\% & 88.34 & 93.75 & 93.75\\
\hline\end{tabular}
\end{table*}

\begin{figure*}[!t]
\centering
\includegraphics[width=\textwidth,height=.33\textheight,keepaspectratio]{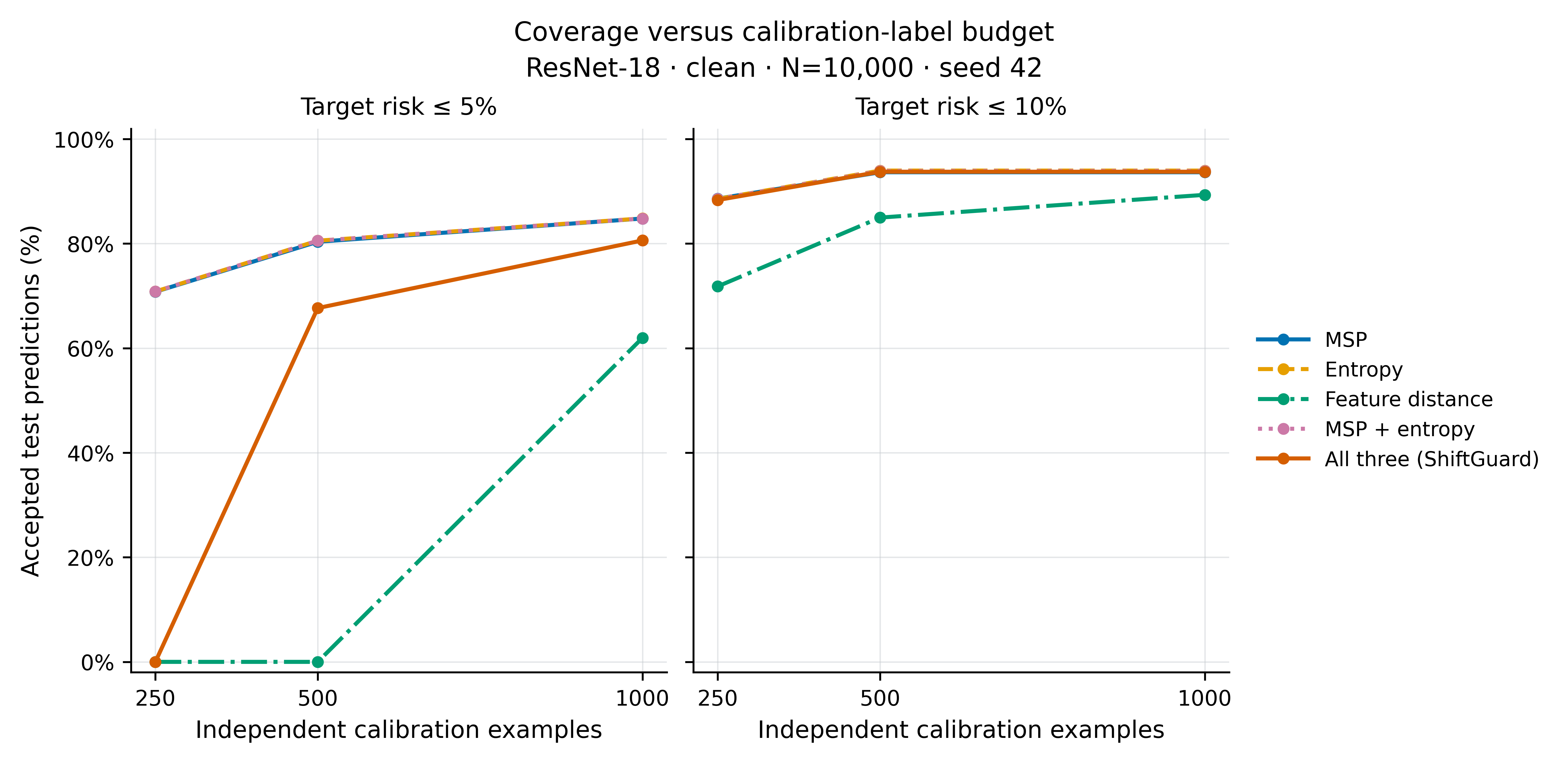}
\caption{Certified-rule test coverage versus nested calibration budget at each risk target. Budgets are dependent prefixes, not repeated experiments. Zero coverage indicates not certified.}\label{fig:budget}
\end{figure*}

More calibration data gave more certified coverage. For entropy at the 5\% target, test coverage rose from 70.84\% with 250 calibration images to 80.56\% with 500 and 84.82\% with 1,000. The extra data tighten the upper bound, so rules that accept more examples become eligible. The two methods that use distance were hit hardest by a small budget. With 250 calibration images, neither distance alone nor the three-signal fusion could certify the 5\% target, so both returned "not certified" and rejected everything, while MSP, entropy and the two-signal fusion each certified about 71\% coverage. With 500 images the three-signal fusion certified only 67.67\% against 80.56\% for entropy. The gap narrowed with 1,000 images (80.61\% against 84.82\%) and did not close. So in this experiment the added complexity did not become useful as calibration data grew. It became less costly.

At the 10\% target, coverage for the confidence methods was the same at 500 and 1,000 images, because the same thresholds were chosen. This is a property of the finite grid. The best eligible grid point stayed the same once enough calibration data were available.

\subsection{Calibration of the error probabilities}

The fused scores output estimated error probabilities, so we also checked them as probabilities. The two-signal fusion had a binary Brier score of 0.0533 and an ECE of 0.0261, and the three-signal fusion had a Brier score of 0.0535 and an ECE of 0.0192. The three-signal score is slightly better calibrated in the ECE sense and equal in Brier score, while ranking errors clearly worse. Ranking and calibration are different properties. A score can give well-scaled probabilities and still order examples badly at the confident end. These ECE values come from 15-bin estimates with a small number of errors, and sparse bins fluctuate, so we do not read much into the 0.007 gap.

\subsection{Sampling uncertainty (post hoc)}

The pipeline reports single-seed point estimates, so we added one analysis after the fact to judge how large a difference would have to be to stand out from test-sample noise. Using the saved test scores, we drew 2,000 paired bootstrap resamples of the 10,000 test images and recomputed error AUROC for every method on each resample. This captures uncertainty from which test images were drawn. It does not capture variability from training, which would need more seeds. It was run on frozen scores and informed no selection or threshold. We disclose it as post hoc.

\begin{table*}[!t]
\centering\footnotesize\setlength{\tabcolsep}{4pt}
\caption{Post hoc paired AUROC differences: 2,000 resamples; 95\% percentile intervals.}
\begin{tabular}{lrr}\hline
Comparison & Difference & Interval\\\hline
Entropy minus MSP & +0.00033 & [-0.00002, +0.00067]\\
MSP+Entropy minus MSP & +0.00054 & [-0.00041, +0.00147]\\
MSP+Entropy minus Entropy & +0.00021 & [-0.00046, +0.00089]\\
Combined minus MSP+Entropy & -0.04209 & [-0.04917, -0.03519]\\
Combined minus MSP & -0.04155 & [-0.04881, -0.03466]\\
Distance minus MSP & -0.10058 & [-0.11014, -0.09104]\\
\hline\end{tabular}
\end{table*}

The intervals for the confidence-only comparisons contain zero and do not resolve their small differences. The deficit of the three-signal fusion relative to the two-signal fusion is about 0.042 and its interval is far from zero, so on this test set the estimated deficit is large relative to the bootstrap variation. The single-seed caveat still applies. The interval indicates that the deficit is large relative to the estimated test-sample variability. It does not establish that the effect would appear for another training run.

\section{Discussion}

\subsection{Why distance did not help}

We did not test why, so what follows are hypotheses that fit the numbers.

The first is redundancy. A well-trained network that has fit its training data nearly perfectly (the final training loss was 0.0068) concentrates its confident predictions on feature vectors close to class centroids, so the information in distance and the information in confidence overlap. The rank correlation between the distance score and either MSP or entropy was about 0.39, which is real but moderate. The part of distance that is not already in confidence may be mostly noise for the purpose of detecting errors.

The second is the fusion objective. A logistic regression fitted to maximise the likelihood of the error labels is not trained to rank, and its output is a weighted sum of standardised features. The MSP signal is very skewed, with mean 0.981 and standard deviation 0.074 on the fitting set, so after standardisation a handful of low-confidence examples have extreme values while the large mass of confident correct examples is packed near one end. Among those, differences in MSP and entropy are tiny in raw units but may still carry rank information about which are likely to be wrong. Adding a distance term of comparable scale lets noise in distance reorder that crowded region. The pattern in Table 5 matches this story: the fused score loses the most at 50\% to 80\% coverage, the confident end of the ranking, and has almost caught up by 90\% coverage. This is a hypothesis about the mechanism. A direct test would fit a rank-based fusion or compare the ordering inside the confident region, which we did not do.

The third is the weakness of the distance signal itself. Our distance is a diagonal, standardised centroid distance. A full-covariance Mahalanobis distance, a Trust Score or a nearest-neighbour distance might be a stronger signal and might behave differently. Our result is about this simple distance, and we do not claim it generalises to the best distance-based methods.

\subsection{Ranking quality and certified coverage move together here}

Confidence-only scores perform strongly both in error ranking and in certified coverage, though small within-group differences need not preserve an exact ordering. Confidence-based scores were at the top on every metric and the distance-based ones were below. Certification adds a second cost for a worse score. A rule that accepts the same number of images with more errors has a higher upper bound and fails the target sooner, which is why distance and the three-signal fusion fell to "not certified" when the budget was small. In a regime where labels are scarce, a noisier score is penalised twice, once in ranking and again in the bound.

\subsection{How to read the certificates}

The accepted-set error rates observed on the test set (about 2\% for a 5\% target) are much lower than the targets. This should not be read as the procedure being miscalibrated. The certificate controls a worst-case bound over a family of 315 simultaneous rules at 95\% confidence, and the Bonferroni correction pushes each individual bound well out. The price of simultaneous validity is conservatism. A valid less conservative design, such as a smaller prespecified family or an appropriate sequential testing scheme, could raise certified coverage; this was not evaluated. We kept a single family covering every displayed method, threshold and budget so that the comparison across methods rests on one common guarantee.

The guarantee applies to a population of clean images from the same source as the calibration data. The calibration images come from the CIFAR-10 training pool and the evaluation images from the official test set. These are nominally the same distribution, but a nominally identical new test set can still shift accuracy \cite{r35}, and nothing in our experiments can verify the IID assumption.

\subsection{What is a finding and what is a hypothesis}

Findings, for this classifier, seed and dataset: the error-detection and AURC numbers in Table 4; the counts in Table 5; the certified coverages in Tables 6 and 7; the failure of the distance-based methods to certify the 5\% target with 250 images; the bootstrap intervals in Table 8.

Hypotheses we did not test: that distance is redundant with confidence for a well-trained network; that the logistic objective and the skew in MSP explain the loss at the confident end; that a stronger distance signal would change the result; that the same ordering would hold across seeds, datasets and architectures.

\section{Limitations}

\begin{enumerate}
\item One training seed. The full ResNet-18 experiment has a single seed (42). The paired bootstrap covers test-sample variability and says nothing about training variability. Differences among MSP, entropy and the two-signal fusion are too small to interpret at this level.

\item One dataset and one architecture. All results come from CIFAR-10 and a CIFAR-adapted ResNet-18 trained from scratch. A pretrained network, a different architecture or a harder dataset might rank the signals differently. Earlier exploratory results described in the draft are not part of the submitted evidence; they are not used to support the main comparison.

\item No shift experiments. The code includes a corruption module, but the corruption list for this run is empty, so there are no corruption results. We make no claim about distribution-shift detection or robustness, and the name ShiftGuard should not be taken as evidence of either. The certificates apply under the stated IID assumption and do not extend to shifted data.

\item A simple distance. The distance signal is a diagonal, standardised distance to the predicted-class centroid, computed from penultimate features. It ignores feature correlations and multimodal classes. Stronger distance-based detectors exist and were not tested.

\item Fixed estimator and grid. The logistic regression uses a prespecified C = 1 and was not tuned, and the candidate grid has 21 quantile thresholds. Other fusion models, other regularisation and finer grids might change the results.

\item Conservative, nested certification. The simultaneous 315-rule family makes the bounds conservative, and the three budgets are nested prefixes of one calibration set, so they are not independent. Observed test risk is a single-sample estimate.

\item IID assumption. The calibration pool is the CIFAR-10 training split and the evaluation set is the official test split. The guarantee relies on these behaving like draws from one population, which cannot be verified here.

\item Small fitting and calibration sets. The fitting set has 76 errors and the calibration set 81, so every fitted coefficient and bound rests on a small number of mistakes. This is the intended low-label regime but it limits precision.

\item Post hoc analysis. The bootstrap in Section 5.7 was added after the main run. It used frozen scores and informed no choices, but it was not prespecified.
\end{enumerate}

\section{Conclusion And Future Work}

We asked whether a feature-distance signal improves on confidence and entropy for detecting classification errors and for accepting predictions at a controlled error rate. On a from-scratch CIFAR-10 ResNet-18 with 92.92\% test accuracy, the answer in this run is no. MSP, entropy and their logistic fusion ranked errors almost identically (AUROC 0.9230, 0.9233 and 0.9235), while distance alone (0.8224) and the three-signal fusion (0.8814) were worse. With 1,000 labelled calibration images and a 5\% risk target, entropy certified a rule that accepted 84.82\% of test images at an observed error of 2.05\%, against 80.61\% at 2.21\% for the three-signal fusion. With 250 calibration images the distance-based methods could not certify the 5\% target at all. A small calibration set made the added complexity a liability and a larger one reduced the cost without removing it.

The strongest supported statement is narrow. For this classifier and seed, a simple standardised centroid distance did not earn its place beside confidence and entropy, and a more complex score made certification harder when labels were scarce. We do not claim that distance signals are useless, that the result holds for other models or data, or that these scores detect distribution shift.

\newpage
Several directions follow. The most useful is more training seeds, so that the comparison has a real across-run variance, and more architectures and datasets. Stronger distance signals, including full-covariance Mahalanobis distance, Trust Score and nearest-neighbour distances, should be tested under the same certification protocol. Rank-based fusion, or fusion that treats the confident region more carefully, would test the explanation in Section 6.1. Finally, the corruption module already in the code can support a shift evaluation with frozen clean thresholds, which would tell us whether certificates degrade under corruption and whether distance helps there, in a separately evaluated follow-up. Any change motivated by the test results here should be evaluated on fresh data and reported as a follow-up analysis, with no tuning on the test set.

\section*{Code and Data Availability}

The author-provided repository is \url{https://github.com/mitravarun123/Shift_Name-}. Public availability was not independently confirmed during preparation. The accompanying research archive preserves the submitted result folder, including configuration, split indices, saved predictions, estimator metadata, certificates and training history. It contains no classifier checkpoint or training source. The manuscript package contains all figures, typeset tables, CSV data and the independent score/certificate audit and bootstrap script. Reproduction of training requires the source repository and checkpoint or a fresh run; numerical reproduction from the saved scores is documented in the package README. No project licence is asserted.
\clearpage\onecolumn\appendices
\section{Complete Experimental Configuration and Data Integrity}
The main experiment is \texttt{risk\_control\_v1}, architecture \texttt{resnet18}, scope \texttt{full}, seed 42, device CUDA, batch size 128, 40 epochs, initial learning rate 0.001, minimum learning rate 0.00001 and weight decay 0.0001. The run uses two CPU threads, zero data-loader workers, no training/test limit and no corruptions. Feature dimension is 512. Saved figures are 600-DPI PNG exports. Classifier training completed all 40 epochs; the best checkpoint is epoch 39 with selection accuracy 0.9395. Final-epoch selection accuracy is 0.9375 and training loss is 0.0068. Training took approximately 2,837 seconds. Peak RSS is 1,927.91 MiB and peak CUDA allocation 876.59 MiB.

The recorded session specifies Python 3.13.16, PyTorch 2.11.0+cu130, torchvision 0.26.0+cu130, CUDA 13.0, NumPy 2.1.3, scikit-learn 1.6.1, SciPy 1.16.3, Matplotlib 3.10.0 and PyYAML 6.0.3, on a Tesla T4. These are historical run metadata, not the environment used to prepare the manuscript.

The dataset archive is recorded as CIFAR-10 Python format, mirror \path{MIT-OL-AI-D/cifar-10-python}, revision \path{a48007227f9e2cd0af96175f4afb5ac4965e261b}, with archive MD5 \path{c58f30108f718f92721af3b95e74349a}. The saved index arrays independently verify that training, selection, reference, fitting and calibration are disjoint and partition all 50,000 training indices. Training class counts range from 4,483 to 4,516, selection from 199 to 201, reference and fitting from 99 to 101, and calibration from 83 to 118. Test indices belong to the distinct official test pool.

\section{Complete Risk-Certification Audit}
Every row below uses the common confidence family of 315 rules and $\delta=0.05$. A failed certificate has zero coverage and undefined conditional risk (shown as ---). Columns distinguish calibration accepted/error counts, upper risk $U$, test accepted/error counts, test coverage and observed test risk. No test outcome enters certificate selection.

\begingroup\footnotesize\setlength{\tabcolsep}{4pt}\begin{longtable}{llrrrrrr}
\caption{All 30 prespecified method/budget/target selections.}\\
\hline Budget & Method & Target & Cal. accepted/errors & $U$ & Test accepted/errors & Coverage & Test risk\\\hline\endfirsthead
\hline Budget & Method & Target & Cal. accepted/errors & $U$ & Test accepted/errors & Coverage & Test risk\\\hline\endhead
250 & MSP & 5\% & 178/0 & 4.80\% & 7079/46 & 70.79\% & 0.65\%\\
250 & MSP & 10\% & 226/7 & 9.54\% & 8861/244 & 88.61\% & 2.75\%\\
250 & Entropy & 5\% & 178/0 & 4.80\% & 7084/46 & 70.84\% & 0.65\%\\
250 & Entropy & 10\% & 225/6 & 8.89\% & 8856/243 & 88.56\% & 2.74\%\\
250 & Distance & 5\% & --- & --- & 0/0 & 0.00\% & ---\\
250 & Distance & 10\% & 185/3 & 8.03\% & 7182/179 & 71.82\% & 2.49\%\\
250 & MSP + entropy & 5\% & 178/0 & 4.80\% & 7084/46 & 70.84\% & 0.65\%\\
250 & MSP + entropy & 10\% & 225/6 & 8.89\% & 8854/245 & 88.54\% & 2.77\%\\
250 & All three & 5\% & --- & --- & 0/0 & 0.00\% & ---\\
250 & All three & 10\% & 224/7 & 9.62\% & 8834/263 & 88.34\% & 2.98\%\\
500 & MSP & 5\% & 401/4 & 4.22\% & 8036/103 & 80.36\% & 1.28\%\\
500 & MSP & 10\% & 469/22 & 9.23\% & 9365/408 & 93.65\% & 4.36\%\\
500 & Entropy & 5\% & 401/4 & 4.22\% & 8056/105 & 80.56\% & 1.30\%\\
500 & Entropy & 10\% & 468/22 & 9.25\% & 9397/418 & 93.97\% & 4.45\%\\
500 & Distance & 5\% & --- & --- & 0/0 & 0.00\% & ---\\
500 & Distance & 10\% & 416/18 & 9.09\% & 8498/355 & 84.98\% & 4.18\%\\
500 & MSP + entropy & 5\% & 401/4 & 4.22\% & 8055/105 & 80.55\% & 1.30\%\\
500 & MSP + entropy & 10\% & 468/22 & 9.25\% & 9388/407 & 93.88\% & 4.34\%\\
500 & All three & 5\% & 345/4 & 4.89\% & 6767/103 & 67.67\% & 1.52\%\\
500 & All three & 10\% & 466/22 & 9.29\% & 9375/409 & 93.75\% & 4.36\%\\
1000 & MSP & 5\% & 845/15 & 4.04\% & 8480/175 & 84.80\% & 2.06\%\\
1000 & MSP & 10\% & 935/47 & 8.10\% & 9365/408 & 93.65\% & 4.36\%\\
1000 & Entropy & 5\% & 846/15 & 4.03\% & 8482/174 & 84.82\% & 2.05\%\\
1000 & Entropy & 10\% & 938/50 & 8.47\% & 9397/418 & 93.97\% & 4.45\%\\
1000 & Distance & 5\% & 622/11 & 4.53\% & 6196/102 & 61.96\% & 1.65\%\\
1000 & Distance & 10\% & 887/57 & 9.90\% & 8933/445 & 89.33\% & 4.98\%\\
1000 & MSP + entropy & 5\% & 846/15 & 4.03\% & 8479/174 & 84.79\% & 2.05\%\\
1000 & MSP + entropy & 10\% & 936/49 & 8.36\% & 9388/407 & 93.88\% & 4.34\%\\
1000 & All three & 5\% & 799/17 & 4.62\% & 8061/178 & 80.61\% & 2.21\%\\
1000 & All three & 10\% & 929/47 & 8.16\% & 9375/409 & 93.75\% & 4.36\%\\
\hline\end{longtable}\endgroup

\section{Mathematical Details and Validity of Selection}
For fixed scoring and threshold functions, define population acceptance $A_\tau=\mathbf1\{s(X)\le\tau\}$, coverage $C_\tau=\mathbb P(A_\tau=1)$ and conditional risk $r_\tau=\mathbb P(\widehat Y\ne Y\mid A_\tau=1)$ when $C_\tau>0$. Conditional on the accepted calibration count $M=m$, the accepted error count is binomial with parameters $(m,r_\tau)$ under the stated IID assumption. Thus a one-sided Clopper--Pearson upper bound has failure probability at most $\delta'$ for each fixed rule. Conditioning on the independent training/reference/fitting data fixes the score and grid. The union bound gives
\begin{equation}
\mathbb P\!\left(\exists f\in\mathcal F:r_f>U_f\right)\le\sum_{f\in\mathcal F}\delta'=|\mathcal F|\delta'=0.05.
\end{equation}
No independence among candidate tests is required: nested budgets and shared examples do not invalidate this union bound. On the simultaneous event, any data-selected rule satisfying $U_f\le\alpha$ has population conditional risk at most $\alpha$. Reusing the same bounds for two risk targets introduces no additional failure event. These statements provide no guarantee under arbitrary distribution shift and do not bound a particular finite test-set error count.

With zero accepted errors, the exact bound simplifies to
\begin{equation}
U(0,m,\delta')=1-(\delta')^{1/m},\qquad m\ge1.
\end{equation}
At $\delta'=0.05/315$, at least 171 zero-error accepted calibration examples are needed to certify a 5\% target. This explains the 250-image setting: the confidence-only rules accept 178 calibration examples without error and have upper risk approximately 4.80\%. The minimum-acceptance rule of 20 is therefore weaker than the statistical requirement in this particular zero-error case. With no accepted examples the implemented bound is one and the rule is ineligible; its conditional risk remains undefined.

\subsection{Exact-tie treatment in AURC}
Suppose a score-tied group contains $g$ examples and $b$ errors, preceded by $t$ examples and $a$ errors. Under uniform random ordering inside the tied group, the expected prefix risk after $j$ tied examples is
\begin{equation}
\mathbb E[R_{t+j}]=\frac{a+jb/g}{t+j},\qquad j=1,\ldots,g.
\end{equation}
Summing these expected risks over all groups and dividing by $N$ makes AURC invariant to the arbitrary ordering of exact ties. Fixed-coverage summaries use $\lceil cN\rceil$ accepted examples. They describe the test ranking and do not identify independently certified deployment thresholds.

\subsection{Probability metrics}
For true class $y_i$, softmax probabilities $p_{ik}$, bin counts $n_b$, average confidence $\overline q_b$ and accuracy $\overline a_b$,
\begin{align}
\mathrm{NLL}&=-\frac1N\sum_i\log\max(p_{i,y_i},10^{-15}),\\
\mathrm{Brier}_{\rm multi}&=\frac1N\sum_i\sum_{k=1}^{10}(p_{ik}-\mathbf1\{y_i=k\})^2,\\
\mathrm{ECE}_{15}&=\sum_{b:n_b>0}\frac{n_b}{N}|\overline q_b-\overline a_b|.
\end{align}
ECE uses 15 equal-width bins on $[0,1]$, with the rightmost bin including one. Binary error-probability metrics replace class correctness by error labels and use $(\widehat p_i-e_i)^2$ for the binary Brier score. These calibration metrics do not measure ranking quality.

\section{Additional Diagnostics}
The following figures are supplied clean-test or training diagnostics. They share the main experiment's seed 42 and do not provide independent replications. The confusion matrix gives class recalls of 94.0, 97.1, 90.1, 83.7, 94.2, 89.6, 95.2, 94.9, 95.2 and 95.2\% in the standard CIFAR-10 class order (airplane through truck). The largest off-diagonal counts are dog-to-cat 67 and cat-to-dog 66.

\begin{figure}[!t]
\centering
\includegraphics[width=\textwidth,height=.43\textheight,keepaspectratio]{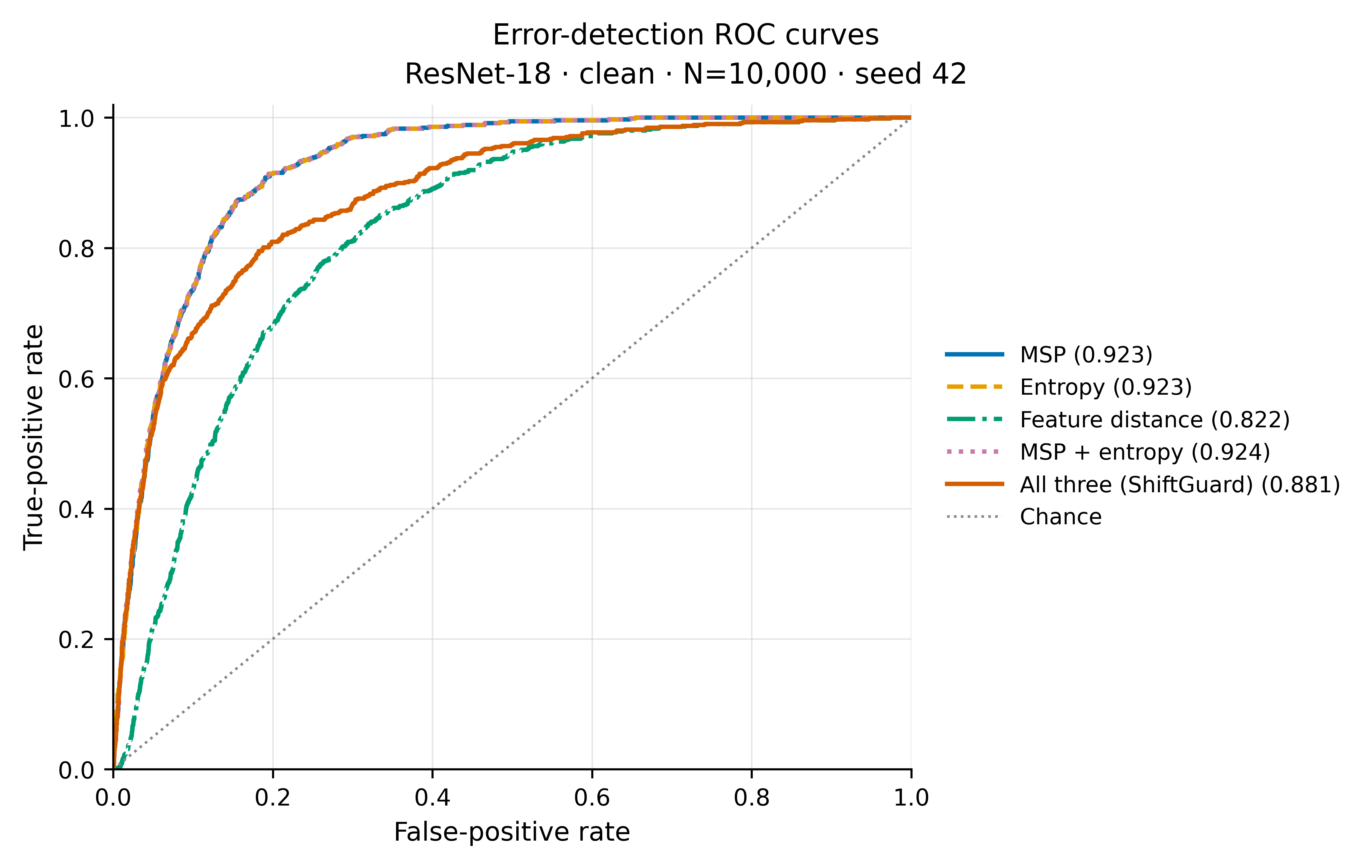}
\caption{Clean-test error-detection ROC curves, with incorrect predictions as positives. One training seed.}\label{fig:roc}
\end{figure}

\begin{figure}[!t]
\centering
\includegraphics[width=\textwidth,height=.43\textheight,keepaspectratio]{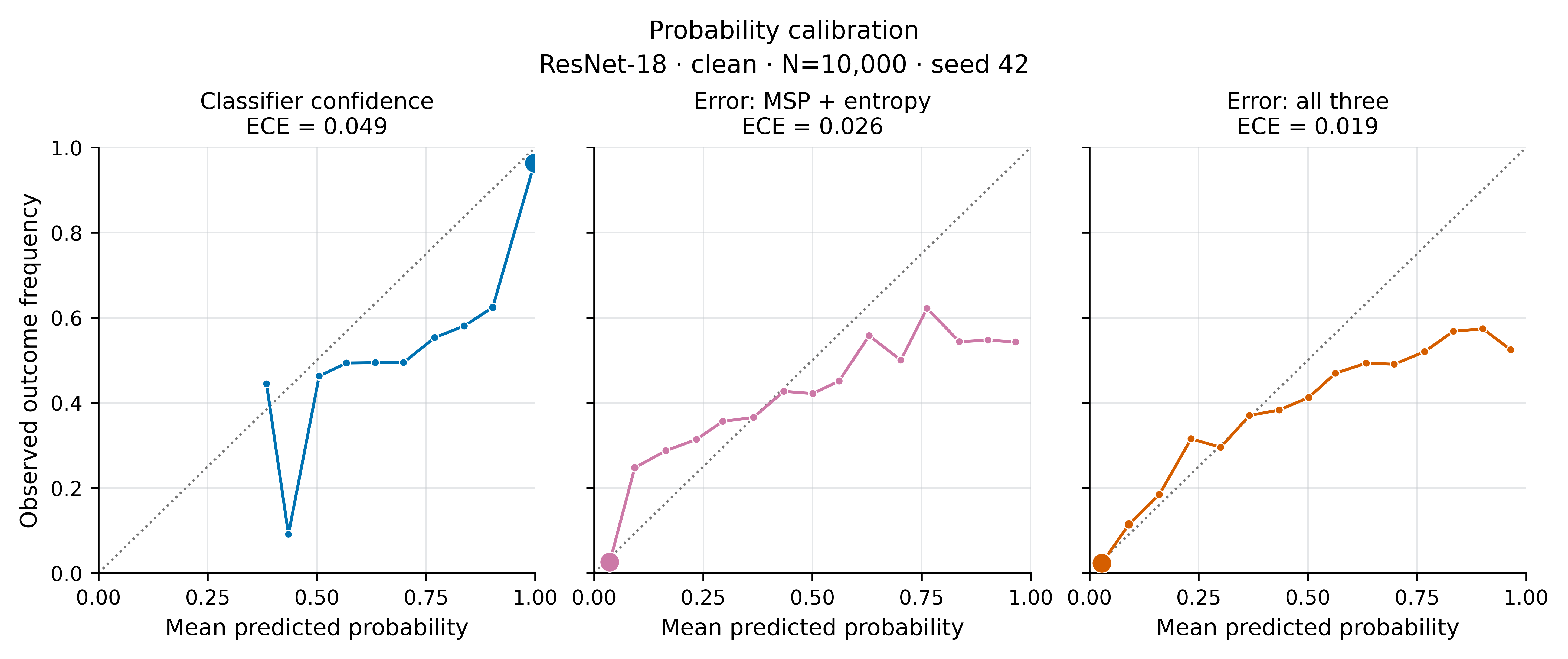}
\caption{Reliability diagrams for classifier correctness and fused error probabilities. Fifteen equal-width bins; marker sizes reflect bin counts. Sparse bins can fluctuate.}\label{fig:calibration}
\end{figure}

\begin{figure}[!t]
\centering
\includegraphics[width=\textwidth,height=.43\textheight,keepaspectratio]{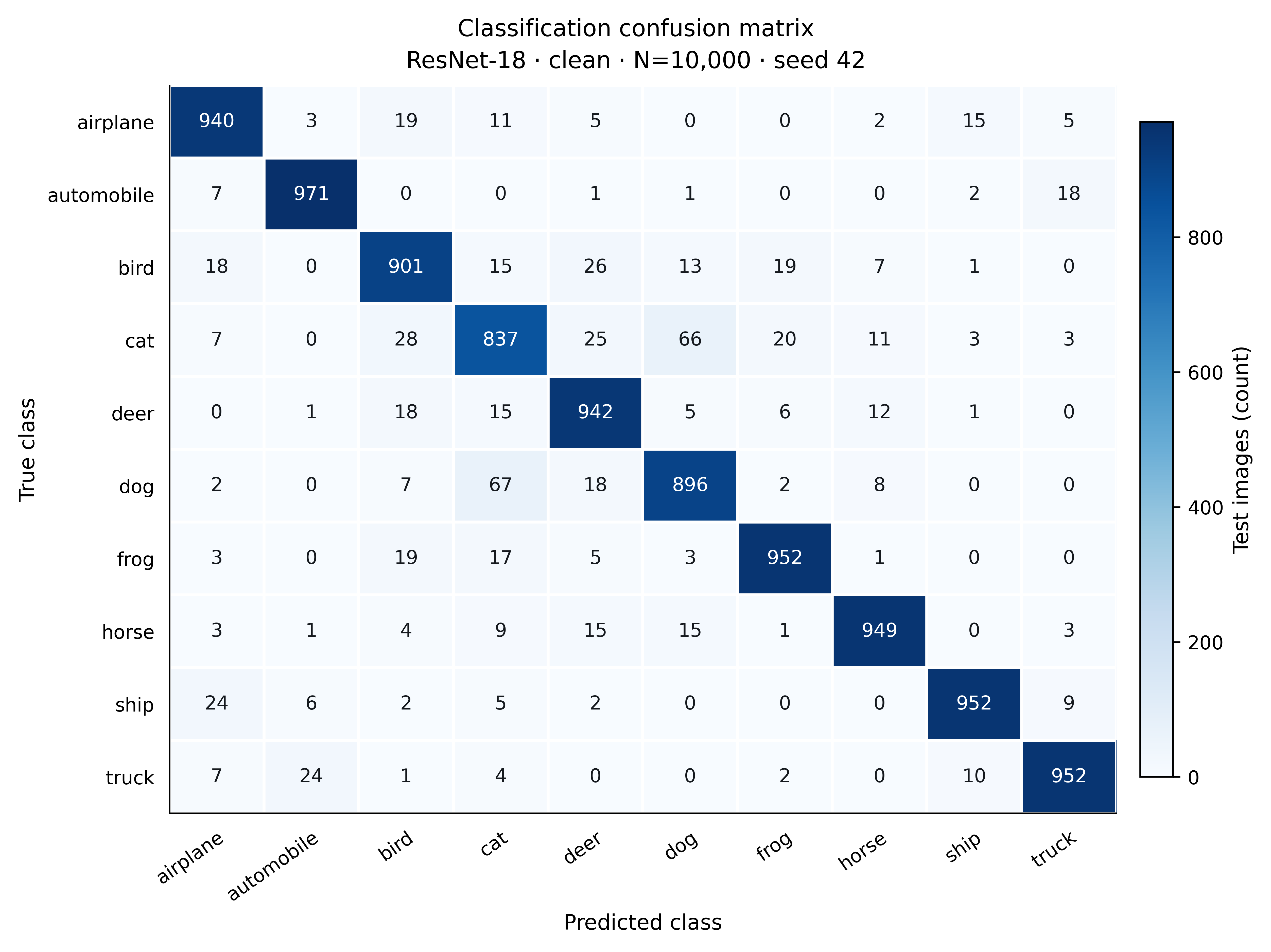}
\caption{Official CIFAR-10 test confusion matrix; entries are image counts.}\label{fig:confusion}
\end{figure}

\begin{figure}[!t]
\centering
\includegraphics[width=\textwidth,height=.43\textheight,keepaspectratio]{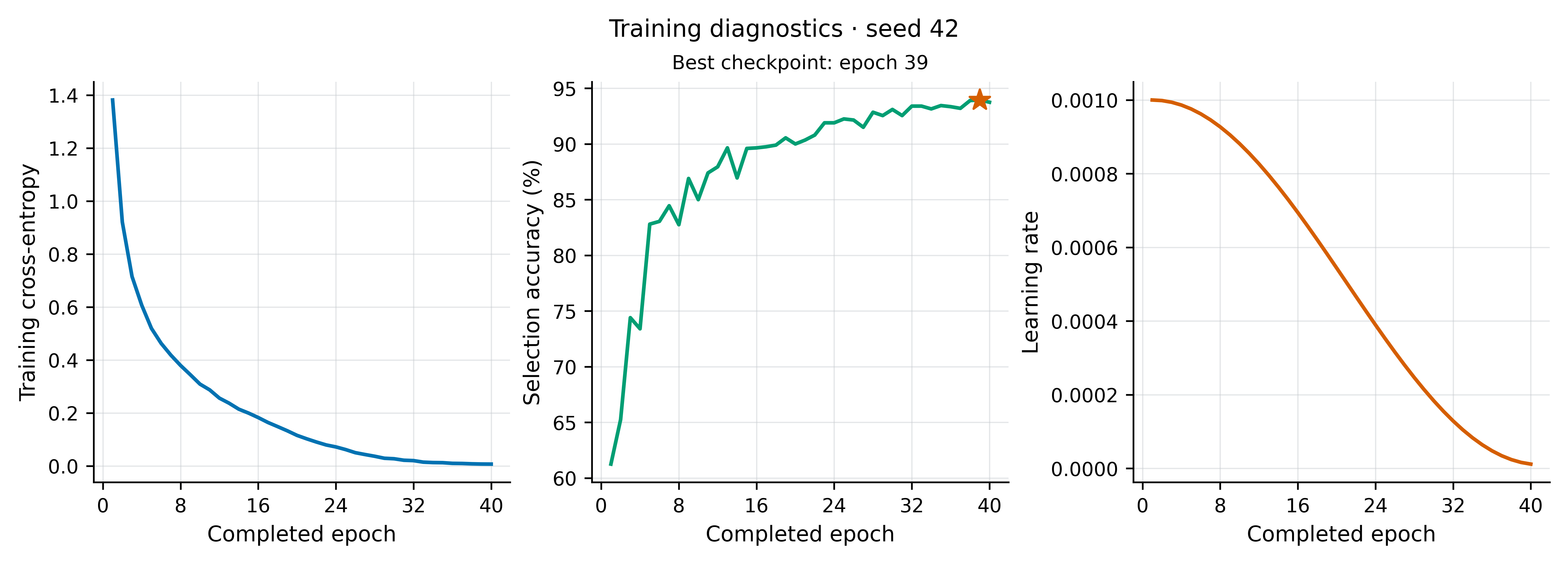}
\caption{Training loss, checkpoint-selection accuracy and learning rate. Epoch 39 is selected by the independent selection subset, not test results.}\label{fig:training}
\end{figure}

\clearpage
\section{Estimator Audit and Post Hoc Uncertainty}
The saved first fusion input equals maximum softmax probability, not its complement. Fitting means are $(0.980919,0.048832)$ and scales $(0.074233,0.158986)$ for the two-signal model. Adding distance appends mean 0.574360 and scale 0.138404. Standardised coefficients are $(0.835964,1.803954)$ with intercept $-2.977201$, and $(0.767487,1.557528,0.587546)$ with intercept $-3.145717$ for the three-signal model. Reconstructing both error probabilities from the saved input matrix reproduces the stored scores to floating-point precision. Estimator pickle/joblib files are preserved as evidence but are not executed by the audit script.

The paired bootstrap was recomputed for manuscript preparation from frozen test scores: 2,000 resamples of 10,000 image indices with replacement, NumPy default random generator seed 0, shared indices across methods, and 2.5th/97.5th percentiles of paired AUROC differences. It is explicitly post hoc, conditional on this trained network, and does not represent training-seed variability. The bootstrap intervals are not adjusted for simultaneous inference over the six displayed contrasts. The accompanying CSV contains the recomputed intervals and the script regenerates them.

\section{Evidence, Reproducibility and Scope}
The independent audit verifies all 315 candidate upper bounds against the exported counts, every one of the 30 selected method/budget/target rules, their test accepted/error counts, all five AUROCs, disjoint training-pool partitions, and both logistic scoring functions. The archive records checkpoint SHA-256 \path{daa75ae61d0d41844ceafcf906e85391a5befcec326e9f71b64b5508ce24b379} and configuration SHA-256 \path{a486c2cb090d86d412f13c4a0e5ae2089c593554315fc947e713c10da9e4259e}. These identifiers are reported metadata; the classifier checkpoint itself is absent, so its hash cannot be independently recomputed from this archive.

The source draft also describes earlier small-CNN runs, a two-epoch pilot, and a 20-test software suite. Their underlying results and test logs are absent from the submitted archive. They are excluded from this paper's quantitative evidence rather than treated as verified additional experiments. All empirical conclusions concern the supplied 40-epoch ResNet-18 run. Training architecture, augmentation and optimiser details beyond the saved configuration are author-reported protocol descriptions, not an independent audit of unavailable training source.

The raw 315-row candidate table is supplied as \path{data/calibration_candidate_bounds.csv}; the full selection audit is \path{data/clean_risk_control.csv}. To reproduce numerical checks and bootstrap intervals, install NumPy, pandas and SciPy and run \texttt{python scripts/verify\_and\_bootstrap.py results} in the research archive. This recomputes evaluation summaries from saved outputs without training a model. Full training reproduction requires the training source, dataset and checkpoint workflow. No training was performed in preparing this manuscript.
\clearpage\twocolumn
\IEEEtriggeratref{20}

\end{document}